\documentclass[conference]{IEEEtran}
\IEEEoverridecommandlockouts
\usepackage{balance}
\usepackage{cite}
\usepackage{amsmath,amssymb,amsfonts}
\usepackage{algorithmic}
\usepackage{graphicx}
\usepackage{textcomp}
\usepackage{xcolor}
\usepackage{threeparttable}
\usepackage{stmaryrd}
\usepackage{gensymb}
\usepackage{multirow}

\def\BibTeX{{\rm B\kern-.05em{\sc i\kern-.025em b}\kern-.08em
    T\kern-.1667em\lower.7ex\hbox{E}\kern-.125emX}}
\begin{document}

\title{Real-time Local Feature with Global Visual Information Enhancement\\
\thanks{This research was funded by National Natural Science Foundation of China (No.61603020, No. U1909215), the Fundamental Research Funds for the Central Universities (No.YWF-20-BJ-J-923), the Key Research and Development Program of Zhejiang Province (No. 2021C03050), and the Scientific Research Project of Agriculture and Social Development of Hangzhou (No. 2020ZDSJ0881)}
\thanks{*Haosong Yue is the corresponding author.}
}

\author{\IEEEauthorblockN{Jinyu Miao\textsuperscript{1,2,3}, Haosong Yue\textsuperscript{2,*}, Zhong Liu\textsuperscript{2}, Xingming Wu\textsuperscript{1,2}, Zaojun Fang\textsuperscript{4}, Guilin Yang\textsuperscript{4}}
\IEEEauthorblockA{\textsuperscript{1} \textit{Hangzhou Innovation Institute, Beihang University,} {Hangzhou, China}\\
\textsuperscript{2} \textit{School of Automation Science and Electrical Engineering, Beihang University,} Beijing, China \\
\textsuperscript{3} \textit{School of Vehicle and Mobility, Tsinghua University,} Beijing, China \\
\textsuperscript{4} \textit{Ningbo Institute of Materials Technology and Engineering, Chinese Academy of Sciences,} Ningbo, China \\
Email: jinyu.miao97@gmail.com, yuehaosong@buaa.edu.cn}
}

\maketitle

\begin{abstract}

Local feature provides compact and invariant image representation for various visual tasks. Current deep learning-based local feature algorithms always utilize convolution neural network (CNN) architecture with limited receptive field. Besides, even with high-performance GPU devices, the computational efficiency of local features cannot be satisfactory. In this paper, we tackle such problems by proposing a CNN-based local feature algorithm. The proposed method introduces a global enhancement module to fuse global visual clues in a light-weight network, and then optimizes the network by novel deep reinforcement learning scheme from the perspective of local feature matching task. Experiments on the public benchmarks demonstrate that the proposal can achieve considerable robustness against visual interference and meanwhile run in real time.

\end{abstract}

\begin{IEEEkeywords}

local feature, feature extraction, feature matching, deep learning

\end{IEEEkeywords}

\section{Introduction}
\label{intro}

Within a few decades that the computer vision algorithms develop, the feature algorithms have arisen and played an important role in various kinds of visual tasks, \textit{e.g.}, simultaneous localization and mapping (SLAM) \cite{orbslam3, slamreview}, structure from motion (SfM) \cite{colmap}, and image stitching. Given that image intensities are vulnerable to visual interference in realistic scenes and dense pixel-level visual information will cost unbearable computational burden, researchers prefer to extract features from images and regard them as a kind of compact and robust image representation.

According to the scope of description, traditional feature algorithm can be roughly categorized into two classes: global feature and local feature. 
Global feature converts the global visual information of the whole image into a single vector or matrix. For example, bag of words (BoW) \cite{bow} and vector of locally aggregated descriptors (VLAD) \cite{vlad} imagine the image as an unordered list of visual words and describe an image by the occurrence of these visual words. And histogram of oriented gradients (HOG) \cite{hog} describe the image by the gradient distribution of pixel intensities. 
Oppositely, local feature detects massive salient pixels in an image (denoted as keypoints), and describes the neighboring area of the keypoint using a high-dimensional vector (denoted as descriptor). The early local feature methods are built upon the human prior knowledge about the quality of pixels. As a milestone of local feature algorithms, scale-invariant feature transform (SIFT) \cite{sift} builds scale space of images and detects extrema in the difference-of-Gaussian function. SIFT \cite{sift} achieves excellent robustness ever since its emergence, but its execution speed is low. For more efficient computation, speeded up robust features (SURF) \cite{surf} and oriented FAST and rotated BRIEF (ORB) \cite{orb} are proposed and popularly utilized in real-time visual applications \cite{fabmap,orbslam3}, but they somewhat sacrifice robustness. 

Motivated by the great success of deep convolution neural network (CNN), many attempts on deep learning-based feature algorithms have been made in recent years. As for global feature, NetVLAD \cite{netvlad} has been proven to be a successful deep CNN-based variant of traditional VLAD algorithm \cite{vlad}. It proposes a generalized VLAD layer with learnable parameters and optimizes the model by metric learning scheme. And in the area of local feature, researchers aim to make CNN model to automatically learn how to detect 
repeatable keypoints and generate invariant high-dimensional descriptors. SuperPoint \cite{superpoint} utilizes a shared encoder and two heads to detect keypoints and extract descriptors, respectively. SuperPoint \cite{superpoint} is firstly enforced to regress ground-truth synthetic corners, and then learn to detect keypoints in two images generated by the homographic adaptation while optimize its descriptors. After that, the training datasets with posed images and 3D map are constantly proposed, the correspondence between different views of the scene can be available. Thus, D2-Net \cite{d2net}, R2D2 \cite{r2d2}, and DISK \cite{disk} optimize their CNN model by apply metric learning between two co-visible images or patches and achieve better performance. And researchers apply attention mechanism in CNN model to further enhance the representative ability of local features \cite{delf,mda}.
These newly emerged methods utilize task-specific optimization scheme without the need of engineer heuristics, so their performances and generalization abilities have been improved. 

\begin{figure}
    \centering
    \includegraphics[width=0.97\linewidth]{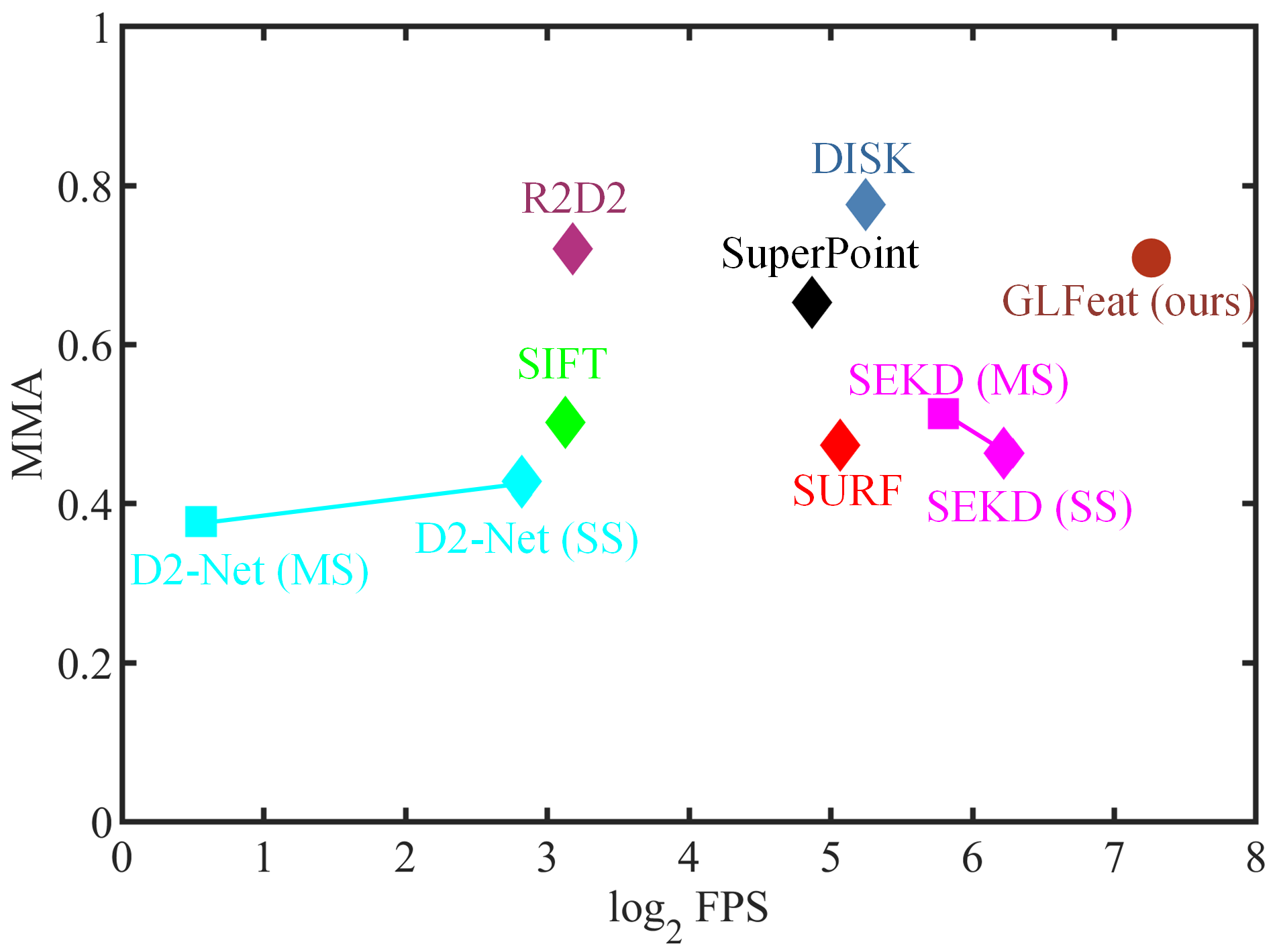}
    \caption{The processed frame per second (FPS) and mean matching accuracy (MMA) on HPatches dataset \cite{hpatches}. The FPS metric is evaluated on a NVIDIA RTX GPU. Detailed setups can be found in Section \ref{fm criteria}.}
    \label{fig:time-mma}
\end{figure}

Although deep CNN-based local features have gradually replaced its traditional counterparts in various visual tasks, some problems still remain. Firstly, it can be seen that the description scope (also called receptive field in deep CNN) of local features are limited in their keypoint detection and descriptor calculation, which make the relevant information for feature extraction is unreliable and sided. The global feature has global receptive field but it cannot detect keypoints, resulting in the limited usage. Besides, as shown in Fig. \ref{fig:time-mma}, the state-of-the-art local features \cite{superpoint,r2d2,disk,sekd,d2net} cannot meet the requirements of efficiency, which blocks the usage of these algorithms in real-time applications.

Under such a context, in this paper, we put our attention on the design of network in local feature algorithm and propose a local feature method with global receptive field, which is denoted as Global-Local Feature (GLFeat). To conduct efficient computation, GLFeat adopt light-weight CNN architecture and simple calculation process. And we introduce global visual information in its deep layers to enhance the representative ability of the model. Finally, GLFeat is optimized by a recently proposed training scheme based on deep reinforcement learning \cite{disk} for better performance. The main contributions of this paper can be summarized as follows:

\begin{itemize}
    \item A light-wight CNN model with global perceptive fields is proposed in GLFeat algorithm, which can efficiently fuse hierarchical feature map with few parameters.
    \item An epipolar constraint-based feature matches annotation method is involved into the novel reinforcement learning-based scheme for GLFeat model optimization.
    \item The proposed GLFeat can achieve considerable robustness and run in real time. 
\end{itemize}

The remainder of this paper is organized as follows: Section \ref{method} describes the proposed GLFeat network and utilized training scheme in details. Comparative experiments withe some state-of-the-art local feature methods are presented in section \ref{exp}. And section \ref{conclusion} ends with conclusions and future works.

\section{Methodology}
\label{method}

Deep learning-based local feature algorithms generally apply a CNN model to convert an image into a high-dimensional feature map, where keypoints are detected and descriptors are sampled. Such methods optimize the learnable parameters of the network by strengthening the response of repeatable and salient pixels while improving the invariance and distinctiveness of descriptors.
To achieve a better balance between computational efficiency and robustness, in this paper, the proposed GLFeat method utilizes light-weight CNN model (Section \ref{model}) and is optimized by deep reinforcement learning-based training scheme (Section \ref{loss}). Detailed description can be found in the following subsections.

\begin{figure*}[!h]
    \centering
    \includegraphics[]{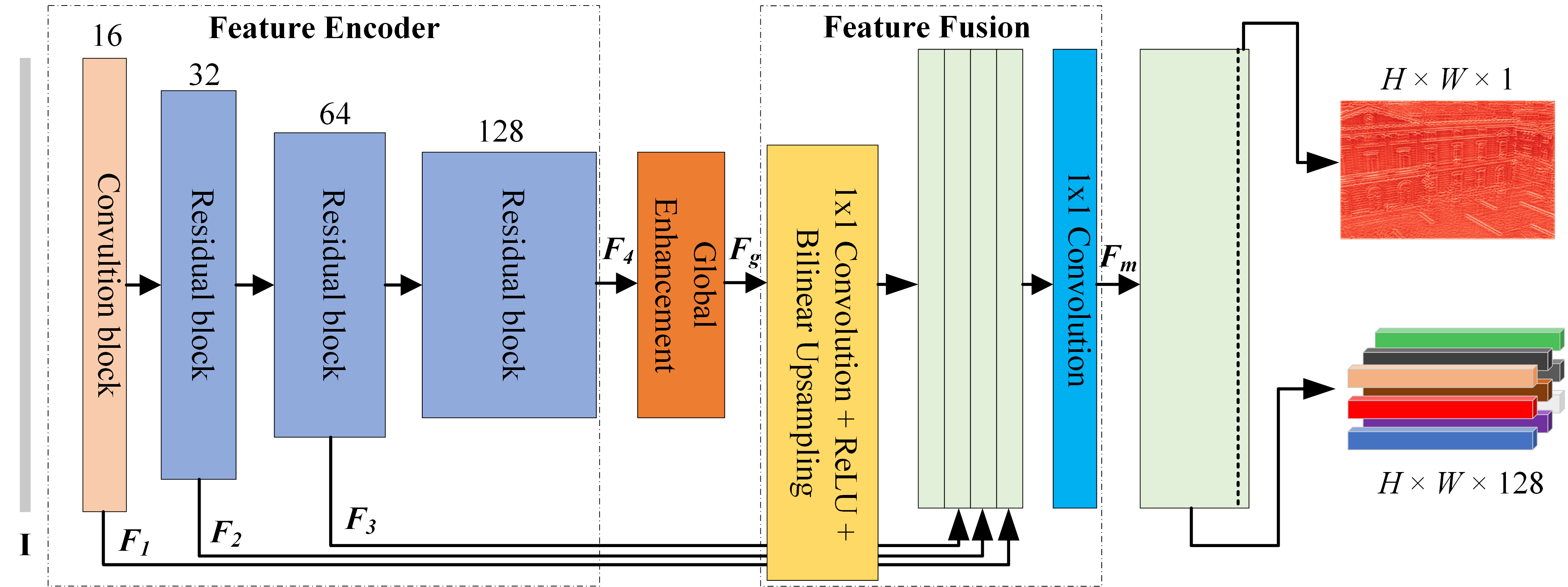}
    \caption{The overall architecture of the proposed GLFeat model.}
    \label{fig:network}
\end{figure*}

\subsection{Network architecture}
\label{model}
In order to effectively build and utilize the feature hierarchy derived by deep CNN, many attempts in local feature algorithms have been made \cite{d2net,r2d2,disk}.
To achieve such a goal with limited parameters, we propose a light-weight CNN architecture, as show in Fig. \ref{fig:network}. The model is feeded by a 3-dimensional RGB images $I \in \mathbb{R}^{H \times W \times 3}$, where $H$ and $W$ are the height and width of the image respectively, and generates four feature maps by a feature encoder module. Then the deepest feature map is enhanced by the global enhancement module and concatenated with the other three feature maps. Finally, a fused 129-dimensional feature map $F_m \in \mathbb{R}^{H \times W \times 129}$ is obtained to extract features.

\textbf{Feature encoder module} (FEM) applies the ALIKE architecture \cite{alike} for its simplicity and lightness. Processed by such a module, the color image is converted into four
hierarchical feature map with different scale and dimension, \textit{i.e.}, $F_1 \in \mathbb{R}^{H \times W \times 16}, F_2 \in \mathbb{R}^{H/2 \times W/2 \times 32}, F_3 \in \mathbb{R}^{H/8 \times W/8 \times 64}, F_4 \in \mathbb{R}^{H/32 \times W/32 \times 128}$. 
The module is composed by a convolution block and three $3 \times 3$ basic residual block \cite{resnet}. The convolution block contains two $3 \times 3$ convolution and ReLU non-linear activation \cite{relu}. Max-pooling process is added before each residual block to change the resolution of feature map.

\begin{figure}[t]
    \centering
    \includegraphics[width=0.97\linewidth]{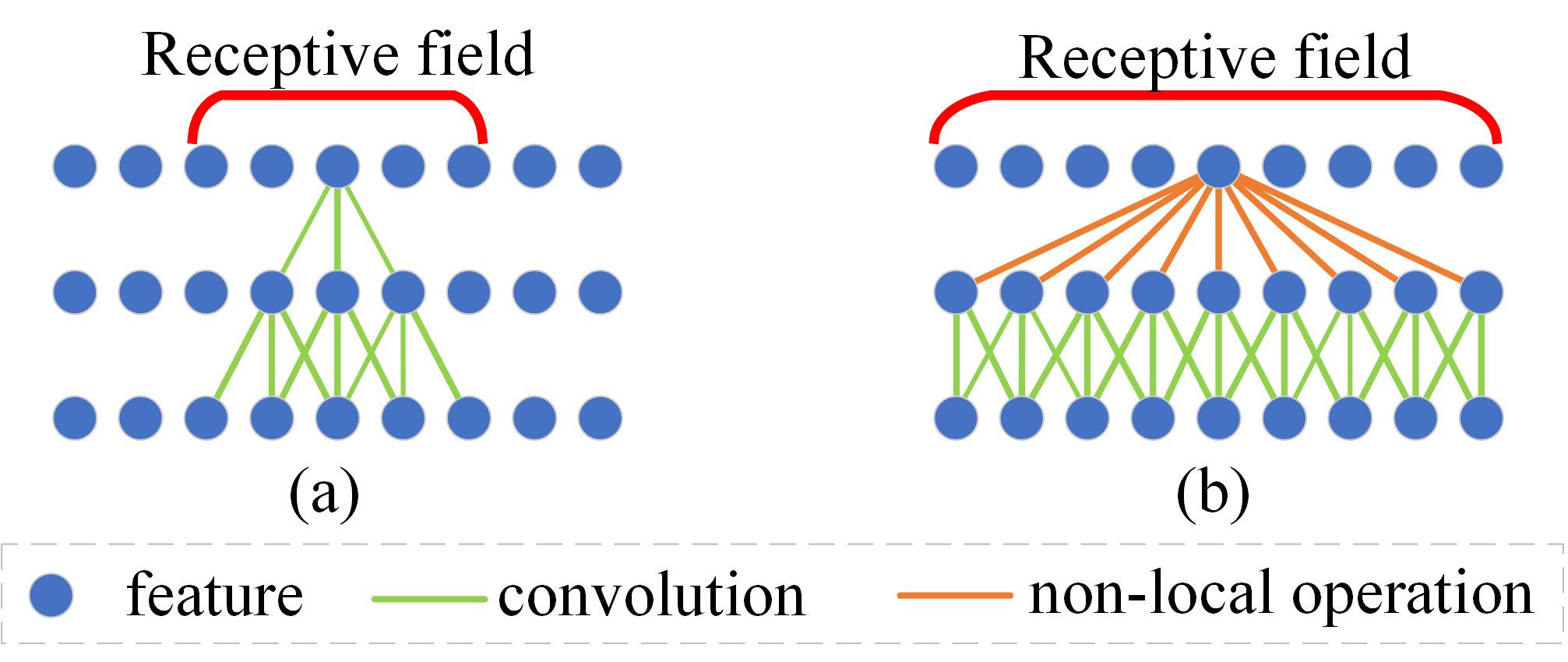}
    \caption{(a) reflects the limited receptive field of traditional CNN model; (b) reflects the global receptive field of CNN model with Non-local operation.}
    \label{fig:nonlocal}
\end{figure}

\textbf{Global enhancement module} (GEM) is proposed to enhance the representative ability of deep feature map in GLFeat. Here, we utilize Non-local block \cite{nonlocal} in this work. In principal, Non-local is technically similar to self-attention mechanism and it enhances the original feature map by weighted summation of the whole feature map, as formulated below:

\begin{equation}
    \label{equ:1}
    y_i = \frac{\sum_{\forall j} f(x_i,x_j)g(x_j)}{\sum_{\forall j} f(x_i,x_j)} + x_i
\end{equation}
where $x_i$ is an original feature map vector and $y_i$ is its enhanced feature map vector. For simplicity, we utilize dot-product formula in the relevance function: $f(x_1,x_2)={\theta(x_1)}^T\phi(x_2)$ and adopt linear embedding formula for $g(\cdot),\theta(\cdot)$, and $\phi(\cdot)$. In this work, we realize the Non-local operation \cite{nonlocal} by the matrix calculation and reduce the scale and dimension of the middle feature map. Specifically, assuming a flatted feature map $X \in \mathbb{R}^{c \times (h \times w)}$, we can define the embedding matrix $W_g, W_\theta, W_\phi \in \mathbb{R}^{\frac{c}{2} \times c}$ in $g(\cdot),\theta(\cdot)$, and $\phi(\cdot)$, respectively. Then, three embeddings can be obtained, \textit{i.e.}, $G=g(X)=W_g X, \Theta=\theta(X)=W_\theta X, \Phi=\phi(X)=W_\phi X$, and the enhanced feature map $Y \in \mathbb{R}^{c \times (h \times w)}$ can be calculated by following formula:

\begin{equation}
    \label{equ:2}
    F_g = {W_x(\mbox{softmax}(\Theta^T\Phi)G^T)}^T + X
\end{equation}
where $W_x \in \mathbb{R}^{c \times \frac{c}{2}}$ lifts the dimension of feature map and $\mbox{softmax}(x_i)=\frac{e^{x_i}}{\sum_i e^{x_i}}$. With the Non-local operation \cite{nonlocal}, the CNN model can obtain global receptive field (as shown in Fig. \ref{fig:nonlocal}), resulting in better representative ability and robustness of feature description.

In this work, a single feature is fused by visual information over the image and it may deteriorate the discriminative ability of feature due to the lack of the location information. Thus, we adopt the 2-dimensional extension of absolute sinusoidal positional embedding \cite{loftr} before Non-local block \cite{nonlocal}:

\begin{equation}
\label{equ:3}
    {PE}^i_{x,y}:= \left\{  
             \begin{array}{rcl}  
                 \mbox{sin}(\omega_k \cdot x)&,& i=4k \\  
                 \mbox{cos}(\omega_k \cdot x)&,& i=4k+1\\  
                 \mbox{sin}(\omega_k \cdot y)&,& i=4k+2\\
                 \mbox{cos}(\omega_k \cdot y)&,& i=4k+3
             \end{array}  
            \right.
\end{equation}
where ${PE}^i_{x,y}$ is the positional embedding for $i^{th}$ channel of the feature on $(x,y)$. The deep feature can be fully enhanced by the global visual information via Non-local operation \cite{nonlocal} and the position information via positional embedding.

\textbf{Feature fusion module} (FFM) aims to adjust the scale and dimension of features and fuse them. In general, the feature map from shallower layer in CNN model will prefer to detailed clues while the deeper feature map focus on abstract semantics. To improve the representative ability of GLFeat, we utilize all the features obtained by the FEM and GEM. By a block composed by a $1\times1$ convolution, a ReLU activation \cite{relu}, and a bilinear upsampling operation, the features $F_i, i\in[1,3]$ and $F_g$ are individually adjusted to the features with the same scale and dimension $F'_* \in \mathbb{R}^{H \times W \times 32}$. These adjusted features are concatenated in the channel dimension and fused by a $1 \times 1$ convolution, resulting in a 129-dimensional feature map $F_m$. The last dimension of $F_m$ is the score map $\mathcal{S} \in \mathbb{R}^{H \times W}$ where keypoints are detected, and others are the dense descriptors $\mathcal{D}$ that need to be L2-normalized later.

\subsection{Training scheme}
\label{loss}

In this work, we choose to utilize a recently proposed training scheme \cite{disk} to optimize our proposed GLFeat model. The scheme adopts deep reinforcement learning technique in network training from the respective of local feature matching. In order to make the non-differentiable local feature extraction and matching process adoptable in the back-propagation procedure, the scheme introduces the probability model of feature extraction and feature matching. 

Following DISK \cite{disk} and SuperPoint \cite{superpoint}, the keypoint heatmap $\mathcal{S}$ is regularly partitioned into cells $u$, whose keypoint heatmap is denoted as $\mathcal{S}^u$. we model the relative salient of a pixel $p$ within the cell $u$ as:
\begin{equation}
\label{equ:4}
    P_s(p|\mathcal{S}^u)={\mbox{softmax}(\mathcal{S}^u)}_p
\end{equation}
which equivalent to the response at $(p)$ of the normalized keypoint heatmap.
Meanwhile, to discard some proposals that located in textureless or unstructured region, we define the absolute quality of each sampled pixel: 
\begin{equation}
\label{equ:5}
    P_a(\mbox{accept}_p|\mathcal{S}^u)=\mbox{sigmoid}(\mathcal{S}^u_p)
\end{equation}
Where $\mbox{sigmoid}(x)=\frac{1}{1+e^{-x}}$. Then, a pixel is sampled and accepted as a keypoint with a probability of $P_s(p|\mathcal{S}^u) \cdot P_a(\mbox{accept}_p|\mathcal{S}^u)$. 
Therefore, using GLFeat, we can detect numerous keypoints and sample corresponding descriptors as $F_I=\{(p_1,D(p_1)),\ldots\}$ from the image $I$ with a probability of $P(F_I|I,\theta_F)$.

In feature matching, we calculate the distance matrix $d$ between features obtained from two images. A cycle-consistent match denotes that the matched features are both closest to each other. To achieve this goal, the $i^{th}$ feature in $I_A$ is matched to the $j^{th}$ fearture in $I_B$ with a probability of $P_{A\rightarrow B}(j|d,i)=\mbox{softmax}(-\theta_M\cdot d(i,:))$ where $\theta_M$ is the hyper-parameter of matching algorithm. Thus, the overall probability of cycle-consistent matching is defined as:

\begin{equation}
\label{equ:6}
    P_{A\leftrightarrow B}(i\leftrightarrow j)=P_{A\rightarrow B}(j|d,i) \cdot P_{B\rightarrow A}(i|d^T,j)
\end{equation}

\begin{figure}[!t]
    \centering
    \includegraphics[width=0.77\linewidth]{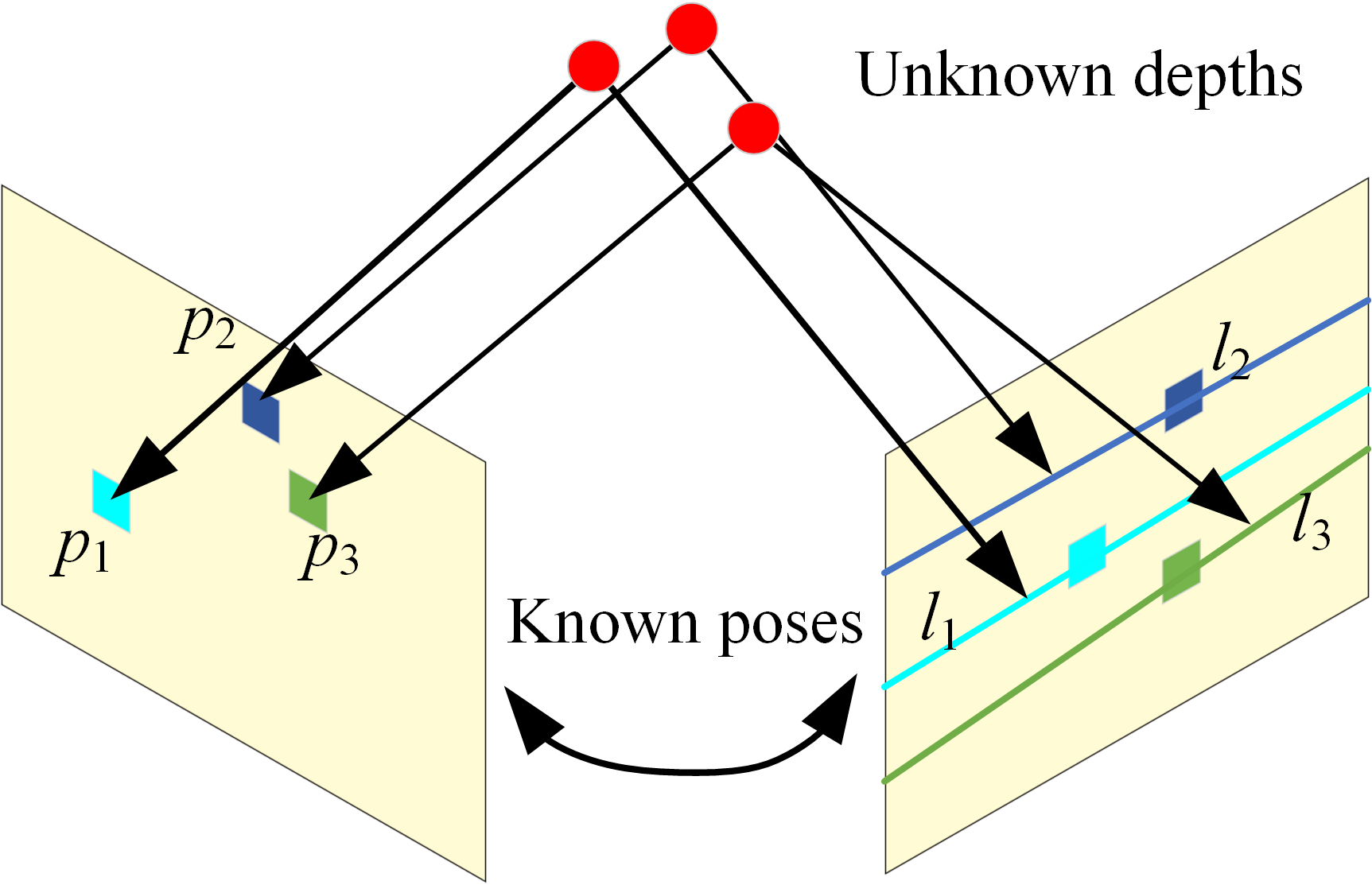}
    \caption{A diagram of epipolar constraint-based feature match annotation.}
    \label{fig:match}
\end{figure}

To make optimization stable and irrespective of the order of matches, the rewards should be individually assigned to each match. We use epipolar constraint to annotate local matches following CASP \cite{casp}, which has lower need of complete depths. As shown in Fig. \ref{fig:match}, a match $(p^i_A,p^j_B) \in M_{A\leftrightarrow B}$ is defined as correct only if it fit the epipolar constraint:

\begin{equation}
\label{equ:7}
    (p^i_A,p^j_B):=\mbox{correct} \Leftrightarrow \mbox{dist}(p^i_A,l^j_B) \le \epsilon \wedge \mbox{dist}(p^j_B,l^i_A) \le \epsilon
\end{equation}
where $\mbox{dist}(p,l)$ is the distance between the point $p$ and the projected epipolar line $l$, $l^j_B,l^i_A$ is the projected epipolar line of $p^j_B,p^i_A$, respectively, and $\epsilon$ is the pre-defined threshold.
    
Built upon the annotation, we give positive rewards $\lambda_{tp}$ to the correct matches, and offer negative penalties $\lambda_{fp}$ to the incorrect matches, and add a small penalties $\lambda_{kp}$ for each sampled keypoint to avoid sampling invalid pixels. And then we estimate the policy gradient \cite{policygradient} to utilize gradient back-propagation algorithm for network training.

\section{Evaluation}
\label{exp}

In this section, we first describe the detailed setups in this work. Then, we evaluate the proposed GLFeat algorithm in the feature matching and visual localization tasks.

\subsection{Experimental setups}
\label{setups}

In this paper, we utilize MegaDepth dataset \cite{megadepth} as the training dataset and sample three covisible images in a triplet to train the model in each step. The training setups are basically similar to DISK \cite{disk} except that we utilize epipolar constraint-based match annotation and add a 5000-steps warmup stage in the first epoch. We also make the parameter $\theta_M$ be a constant and gradually increase it with regular step to avoid ambiguities. Therefore, only the parameters of GLFeat model have been optimized during the training stage. In this work, we train GLFeat for 50 epoches and select the checkpoint with best performance in HPatches benchmark \cite{hpatches}.

During the inference stage, the probability model of feature detection is discarded and we simply apply non-maximum suppression (NMS) on the keypoint heatmap $\mathcal{S}$ to detect keypoints with high response.

For fairness, all the results of compared methods listed in this paper are obtained by the open-sourced codes on the same device with ours. We conduct all the experiments on a server with a NVIDIA RTX GPU.

\subsection{Feature matching}
\label{fm}

To directly evaluate the efficiency and robustness of the proposed GLFeat algorithm, we first conduct the experiments on HPatches benchmark \cite{hpatches} with the respective of feature matching.

\subsubsection{Evaluation criteria}
\label{fm criteria}

108 sequences of HPatches dataset \cite{hpatches} are utilized for evaluation as in \cite{d2net,disk} that each sequence contains a reference image and five test images taken in different illumination or viewpoint conditions. The ground-truth homography matrices between the reference image and test images are also provided. We adopt the public evaluation criteria in feature matching task, \textit{i.e.}, the amount of detected features (NF), repeatability (Rep), matching score (MS), mean matching accuracy (MMA), and mean homography accuracy (MHA). 
We also put our attention on the requirements of computational runtime and memory in the evaluation, so we test the parameters of model (Params) and processed frames per second (FPS) of the method. We extract up to 5000 features per image and only resize images into $640 \times 480$ resolution when testing FPS metric.

\subsubsection{Ablation analysis of global enhancement module}
\label{gib}

\begin{figure}[h]
    \centering
    \includegraphics[width=0.77\linewidth]{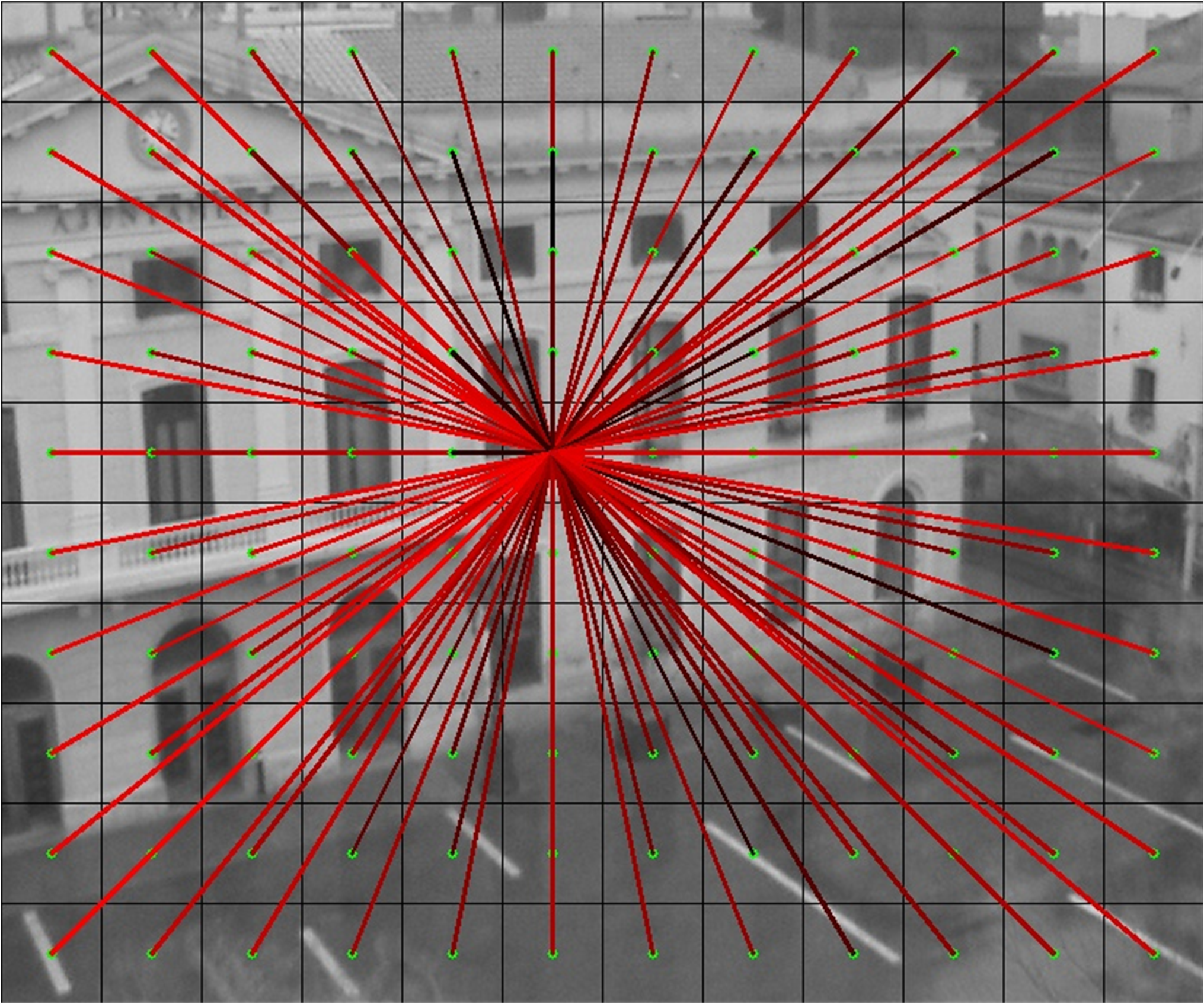}
    \caption{A instance of relevance between one pixel and all the cells in the global enhancement module of GLFeat. A darker line means a stronger relevance.}
    \label{fig:attention}
\end{figure}

In this paper, a global enhancement module (GEM), which is composed by positional embedding block and Non-local block \cite{nonlocal}, is introduced to enhance the representative ability of local features. To conclude the effectiveness of GEM in GLFeat model, we conduct ablation studies of GEM. We train a variant without GEM (``w.o. GEM'') using the same training setups, and compare it with complete GLFeat model (`` w. GEM'') as shown in Table \ref{tab:gem}. The threshold of correct matches is set to 3 pixels. It can be seen that the GEM can effectively improve the robustness of deep CNN model and does not add too much computational and memory burden. As shown in Fig. \ref{fig:attention}, a pixel can obtain visual enhancement from all the cells within the image based on their relevance, thus the representative ability and robustness can be improved.

\begin{table}[!ht]
	\renewcommand{\arraystretch}{1.3}
	\centering
		\caption{Performances of the proposed GLFeat and its variant.}
		\label{tab:gem}
		\begin{tabular*}{\linewidth}{@{}@{\extracolsep{\fill}} l | r r r r r r }
			\hline
			\hline
			Method & Rep & MS & MMA & MHA & Params (M)  \\
			\hline
			w.o. GEM & 0.6099 & 0.4173 & 69.21\% & 70.93\% & 0.3292 \\
			w. GEM  & 0.6178 & 0.4185 & 70.90\% & 71.30\% & 0.3625 \\
			\hline
			\hline
		\end{tabular*}
\end{table}

\subsubsection{Comparison with state-of-the-arts}
\label{fe sota}

To prove the novelty of proposed GLFeat, we compare with some state-of-the-art local feature algorithm, including traditional hand-crafted local features \cite{sift,surf,orb} and recently proposed learned local features \cite{superpoint,disk,d2net,r2d2,sekd}. The hand-crafted local features are implemented using OpenCV 3.4.2 library. The threshold of correct matches in the evaluation is also set to 3 pixels. 
According to the results shown in Table \ref{tab:fm-sota}, one can see that although R2D2 \cite{r2d2} and DISK \cite{disk} achieve excellent performances in matching accuracy and repeatability, their computational speed is low and cannot be adopted in real-time applications such as SLAM. 
The MHA and MS metric of SuperPoint \cite{superpoint} are fine but it cannot extract enough local features due to reduced keypoint heatmap, perhaps leading to unsuccessful initialization and tracking in SLAM applications. 
SEKD \cite{sekd} with multi-scale detection achieves a pleased efficiency and MHA performance, but its repeatability is so bad that it cannot generate enough correct local matches across two images. 
As for the traiditional methods, only SIFT \cite{sift} obtains nice MHA performance but its efficiency is bad without the GPU acceleration.
As a comparison, our proposed GLFeat can achieve a comparable performance in all the criteria and especially, GLFeat adopts a lighter model and simpler calculation, resulting in much faster execution speed and less memory requirement, which denotes a better attempt about the balance between robustness and efficiency.

\begin{table*}[ht]
	\renewcommand{\arraystretch}{1.3}
	\centering
	\begin{threeparttable}
		\caption{Feature matching performances of the proposed GLFeat and the compared methods.}
		\label{tab:fm-sota}
		\begin{tabular*}{\linewidth}{@{}@{\extracolsep{\fill}} l l l | r r r r r r r r }
			\hline
			\hline
			Category & Method & From & NF & Rep & MS & MMA & MHA & Params (M) & FPS  \\
			\hline
			\multirow{2}{*}{hand-crafted} & SIFT \cite{sift} & IJCV'04   & 2890 & 0.4792 & 0.2383 & 50.22\% & \textcolor{blue}{\textbf{72.22\%}} & - & 8.74 \\
			 & SURF \cite{surf} & CVIU'08   & 384 & 0.4162 & 0.2188 & 47.38\% & 46.85\% & - & 33.49 \\
			\hline
			\multirow{7}{*}{learned} & SuperPoint \cite{superpoint} & CVPRW'18   & 1429 & 0.5603 & \textcolor{green}{\textbf{0.4288}} & 65.31\% & \textcolor{green}{\textbf{72.59\%}} & 1.3009 & 29.21 \\
			 & D2-Net (SS) \cite{d2net}      & CVPR'19    & 3843 & 0.3348 & 0.2143 & 42.78\% & 40.19\% & 7.6353 & 7.07 \\
			 & D2-Net (MS) \cite{d2net}      & CVPR'19    & 4414 & 0.3441 & 0.1606 & 37.77\% & 39.44\% & 7.6353 & 1.47 \\
			 & R2D2 \cite{r2d2}             & NeurIPS'19 & \textcolor{red}{\textbf{4922}} & \textcolor{red}{\textbf{0.6437}} & 0.2923 & \textcolor{green}{\textbf{72.09\%}} & 66.85\% & \textcolor{green}{\textbf{0.4844}} & 9.07 \\
			 & SEKD (SS) \cite{sekd}         & arxiv'20        & 4236 & 0.4470 & 0.2650 & 46.32\% & 69.26\% & \textcolor{blue}{\textbf{0.6618}} & \textcolor{green}{\textbf{74.55}} \\
			 & SEKD (MS) \cite{sekd}         & arxiv'20       & \textcolor{blue}{\textbf{4550}} & 0.5226 & 0.2959 & 51.28\% & \textcolor{red}{\textbf{74.07\%}} & \textcolor{blue}{\textbf{0.6618}} & \textcolor{blue}{\textbf{55.45}} \\
			 & DISK \cite{disk}             & NeurIPS'21 & 4423 & \textcolor{green}{\textbf{0.6213}} & \textcolor{red}{\textbf{0.4592}} & \textcolor{red}{\textbf{77.59\%}} & 69.87\% & 1.0924 & 37.95 \\
			\hline
			learned & \multicolumn{2}{c|}{GLFeat (ours)}               & \textcolor{green}{\textbf{4989}} & \textcolor{blue}{\textbf{0.6178}} & \textcolor{blue}{\textbf{0.4185}} & \textcolor{blue}{\textbf{70.90\%}} & 71.30\% & \textcolor{red}{\textbf{0.3625}} & \textcolor{red}{\textbf{153.39}}\\
			\hline
			\hline
		\end{tabular*}
		
		\begin{tablenotes}
			\item The top-3 performances of each criteria are emphasized in \textcolor{red}{\textbf{RED}}, \textcolor{green}{\textbf{GREEN}}, and \textcolor{blue}{\textbf{BLUE}}, respectively. ORB \cite{orb} algorithm achieves extremely bad results so we do not report them here.
		\end{tablenotes}
	\end{threeparttable}
\end{table*}

\subsection{Visual localization}
\label{vloc}

For comprehensive evaluation, we also evaluate the performance of proposed GLFeat when plugged into a public visual localization system\footnote{https://github.com/tsattler/visuallocalizationbenchmark}.

\subsubsection{Evaluation criteria}
\label{vl criteria}

We conduct the experiments on Aachen day-night visual localization benchmark \cite{aachen}. The benchmark registers query images in night time into the 3D model generated by COLMAP \cite{colmap} and recovers the poses of query images. We adopt the public evaluation criteria\footnote{https://www.visuallocalization.net/} that measures the percentage of the query images whose recovered pose is within specific threshold, \textit{i.e.}, ($0.25m, 2\degree$)/($0.5m,5\degree$)/($5m,10\degree$). We still extract up to 5000 features per image and utilize simple mutually matching to individually evaluate the local feature algorithm.

\subsubsection{Comparison with state-of-the-arts}
\label{vloc sota}

We also compare our proposed GLFeat with state-of-the-art local features \cite{sift,superpoint,d2net,r2d2,sekd,disk}.
As shown in Table \ref{tab:vloc}, it can be found that hand-crafted SIFT \cite{sift} performs bad in such a challenging task since it lack robustness against large day-night appearance changes and viewpoint changes. The most of learnt local features can achieve pleased performances. 
D2-Net \cite{d2net} shows much better robustness in visual localization task than its performance in feature matching tasks, which is probably benefit from iterate bundle adjustment. 
Under each tolerance, our proposed GLFeat can achieve a leading or second performance, which ensures a potential wide usage in practical applications. 

\begin{table}[!t]
	\renewcommand{\arraystretch}{1.3}
	\centering
	\begin{threeparttable}
		\caption{Visual localization performances of local feature methods.}
		\label{tab:vloc}
    	\begin{tabular*}{\linewidth}{@{}@{\extracolsep{\fill}}l|rrr}
    		\hline
    		\hline
    		\multirow{2}{*}{Method} & \multicolumn{3}{c}{Percentage of queries} \\
    		& $0.25m, 2\degree$ & $0.5m,5\degree$ & $5m,10\degree$ \\
    		\hline
    		SIFT \cite{sift} & 20.4\% & 27.6\% & 37.8\% \\
    		SuperPoint \cite{superpoint} & 72.4\% & 81.6\% & 93.9\% \\
    		D2-Net (SS) \cite{d2net} & \textcolor{blue}{\textbf{77.6\%}} & \textcolor{red}{\textbf{92.9\%}} & \textcolor{red}{\textbf{100.0\%}}  \\
    		R2D2 \cite{r2d2} & 69.4\% & \textcolor{blue}{\textbf{84.7\%}} & \textcolor{blue}{\textbf{98.0\%}}  \\
    		SEKD (SS) \cite{sekd} & 56.1\% & 69.4\% & 79.6\%  \\
    		SEKD (MS) \cite{sekd} & 67.3\% & 74.5\% & 83.7\%  \\
    		DISK \cite{disk} & \textcolor{red}{\textbf{80.6\%}} & \textcolor{green}{\textbf{86.7\%}} & \textcolor{green}{\textbf{99.0\%}}  \\
    		\hline
    		GLFeat (ours)  & \textcolor{green}{\textbf{79.6\%}} & \textcolor{green}{\textbf{86.7\%}} & \textcolor{red}{\textbf{100.0\%}}  \\
    		\hline
    		\hline
    	\end{tabular*}
    	
    	\begin{tablenotes}
			\item The top-3 performances are emphasized in \textcolor{red}{\textbf{RED}}, \textcolor{green}{\textbf{GREEN}}, and \textcolor{blue}{\textbf{BLUE}}, respectively.
		\end{tablenotes}
		    
	\end{threeparttable}
\end{table}

\section{Conclusion}
\label{conclusion}

In this paper, we make attempts to the problem of local features that the efficiency and robustness cannot be achieved simultaneously. 
We propose a local feature algorithm, denoted as GLFeat, which utilizes a light-weight CNN architecture with a novel global enhancement module (GEM) and is optimized by deep reinforcement learning technique from the view of local feature matching task. 
Experimental evaluation concludes the effectiveness of GEM and the proposed GLFeat can achieve considerable robustness while has a much better computational efficiency and less memory requirements than existing methods. 
It can be widely adopted in various real-time visual applications, such as SLAM and visual (re-)localization.

\bibliographystyle{IEEEtran}
\balance
\bibliography{ref}

\end{document}